\documentclass{article}

    \usepackage[final,nonatbib]{tackling_climate_workshop_style}

\usepackage[utf8]{inputenc} 
\usepackage[T1]{fontenc}    
\usepackage{hyperref}       
\usepackage{url}            
\usepackage{booktabs}       
\usepackage{amsfonts}       
\usepackage{nicefrac}       
\usepackage{microtype}      
\usepackage{multirow}
\usepackage{graphicx}
\usepackage{amsmath}
\usepackage{siunitx}
\usepackage{amssymb}

\title{Don't Waste Data: Transfer Learning to Leverage All Data for Machine-Learnt Climate Model Emulation}

\author{
  Raghul Parthipan$^{1,2}$ \\
  \texttt{rp542@cam.ac.uk} \\
  \And
  Damon J. Wischik$^1$ \\
  \texttt{damon.wischik@cl.cam.ac.uk} \\
  \AND
  \\
  \textsuperscript{1} Department of Computer Science and Technology, University of Cambridge, UK
  \\
   \textsuperscript{2} British Antarctic Survey, Cambridge, UK
   }
\begin{document}

\maketitle

\begin{abstract}

How can we learn from all available data when training machine-learnt climate models, without incurring any extra cost at simulation time? Typically, the training data comprises coarse-grained high-resolution data. But only keeping this coarse-grained data means the rest of the high-resolution data is thrown out. We use a transfer learning approach, which can be applied to a range of machine learning models, to leverage all the high-resolution data. We use three chaotic systems to show it stabilises training, gives improved generalisation performance and results in better forecasting skill. Our code is at \url{https://github.com/raghul-parthipan/dont_waste_data}.


\end{abstract}

\section{Introduction}



Accurate weather and climate models are key to climate science and decision-making. Often we have a high-resolution physics-based model which we trust, and want to use that to create a lower-cost (lower-resolution) emulator of similar accuracy. There has been much work using machine learning (ML) to learn such models from data \cite{arcomano2022hybrid,brenowitz2018prognostic,brenowitz2019spatially,chattopadhyay2020data,chattopadhyay2020sp,gan_hannah,gentine,krasnopolsky,ogorman_2018,rasp2018deep,vlachas2022multiscale,yuval_rf,yuval2021,vlachas2018data}, due to the difficulty in manually specifying them. 

The naive approach is to use coarse-grained high-resolution model data as training data. The high-resolution data is averaged onto the lower-resolution grid and treated as source data. The goal is to match the evolution of the coarse-grained high-resolution model using the lower-resolution one. Such procedures have been used successfully \cite{brenowitz2018prognostic,brenowitz2019spatially,krasnopolsky,yuval_rf,christensen2020constraining,bolton2019applications}. This has several benefits over using observations, including excellent spatio-temporal coverage. But it has a key downside --- the averaging procedure means much high-resolution data is thrown away. 

Our novelty is showing that climate model emulation can be framed as a \textit{transfer learning} task. We can do better by using all of the high-resolution data as an auxiliary task to help learn the low-resolution emulator. And we can do this without any further cost at simulation time. As far as we know, this has not yet been reported in the climate literature. This results in improved generalization performance and forecasting ability, and we demonstrate this on three chaotic dynamical systems.

\paragraph{Related Work.} 

Transfer learning (TL) has been successfully used for fine-tuning models, including sequence models, in various domains such as natural language processing (NLP) and image classification. There are various methods used such as (1) fine-tuning on an auxiliary task and then the target task \cite{phang2018sentence,girshick2014rich,cui2018large}; (2) multi-task learning, where fine-tuning is done on the target task and one or more auxiliary tasks simultaneously \cite{bingel2017identifying,changpinyo2018multi,luong2015multi,peng2020empirical,collobert2008unified,liu2016recurrent,ge2017borrowing}; and mixtures of the two. Our approach is most similar to the first one. However, our models are not pre-trained as is standard in NLP. Despite this, we show our approach remains successful. 

\paragraph{Climate Impact.}

A major source of inaccuracies in weather and climate models arises from `unresolved' processes (such as those relating to convection and clouds) \cite{schneider_clouds,webb2013origins,sherwood2014spread,wilcox2007frequency,ceppi2016clouds,wang2015processes}. These occur at scales smaller than the resolution of the climate model but have key effects on the overall climate. For example, most of the variability in how much global surface temperatures increase after $\textrm{CO}{_2}$ concentrations double is due to the representation of clouds \cite{schneider_clouds,sherwood2014spread,zelinka2020causes}. There will always be processes too costly to be explicitly resolved by our current operational models.

The standard approach to deal with these unresolved processes is to model their effects as a function of the resolved ones. This is known as `parameterization' and there is much ML work on this \cite{arcomano2022hybrid,brenowitz2018prognostic,brenowitz2019spatially,chattopadhyay2020data,chattopadhyay2020sp,gan_hannah,gentine,krasnopolsky,ogorman_2018,rasp2018deep,vlachas2022multiscale,yuval_rf,yuval2021,vlachas2018data}. We propose that by using all available high-resolution data, better ML parameterization schemes and therefore better climate models can be created.

\section{Methods}

Our approach is a two-step process: first, we train our model on the high-resolution data, and second, we fine-tune it on the low-resolution (target) data. 

We denote the low-resolution data at time $t$ as $\mathbf{X}_t \in \mathbb{R}^d$. The goal is to create a sequence model for the evolution of $\mathbf{X}_t$ through time, whilst only tracking $\mathbf{X}_t$. We denote the high-resolution data at time $t$ as $\mathbf{Y}_t\in \mathbb{R}^{dm}$. In parameterization, $\mathbf{X}_t$ is often a temporal and/or spatial averaging of $\mathbf{Y}_t$. We wish to use $\mathbf{Y}_t$ to learn a better model of $\mathbf{X}_t$.

A range of ML models for sequences may be used, but we suggest they should contain both shared and task-specific layers.

We first model $\mathbf{Y}_t$, training in the standard teacher-forcing way for ML sequence models. We use the framework of probability, and so train by maximising the log-likelihood of $\mathbf{Y}_t$, $\log\mathrm{Pr}(\mathbf{y}_1,\mathbf{y}_2,...,\mathbf{y}_n)$. Informally, the likelihood measures how likely $\mathbf{Y}_t$ is to be generated by our sequence model. Next, the weights of the shared layers are frozen and the weights of the target-specific layers are trained to model the low-resolution training data, $\mathbf{X}_t$. Again, under the probability framework, this means maximising the log-likelihood of $\mathbf{X}_t$, $\log\mathrm{Pr}(\mathbf{x}_1,\mathbf{x}_2,...,\mathbf{x}_n)$.

\subsection{RNN Model}

We use the recurrent neural network (RNN) to demonstrate our approach (though it is not limited to the RNN). RNNs are well-suited to parameterization tasks \cite{chattopadhyay2020data,chattopadhyay2020sp,vlachas2022multiscale,vlachas2018data,vlachas2020backpropagation} as they only track a summary representation of the system history, reducing simulation cost. This is unlike the Transformer \cite{vaswani2017attention} which requires a slice of the actual history of $\mathbf{X}_t$. 

For our RNN, the hidden state is shared and its evolution is described by $ \mathbf{h}_{t+1} = f_\theta(\mathbf{h}_t,\mathbf{X}_t)$ where $\mathbf{h}_t \in \mathbb{R}^H$ and $f_\theta$ is a GRU cell \cite{gru}. We model the low-resolution data as
\begin{equation}
    \mathbf{X}_{t+1} = \mathbf{X}_{t} + g_\theta(\mathbf{h}_{t+1}) + \sigma \mathbf{z}_t 
    \label{eq:rnn1}
\end{equation}
and the high-resolution as
\begin{equation}
    \mathbf{Y}_{t+1} = \mathbf{Y}_{t} + j_\theta(\mathbf{h}_{t+1}) + \rho \mathbf{w}_t 
    \label{eq:rnn2}
\end{equation}
where the functions $g_\theta$ and $j_\theta$ are represented by task-specific dense layers, $\mathbf{z}_t \sim \mathcal{N}(0,I)$ and $\mathbf{w}_t \sim \mathcal{N}(0,I)$. The learnable parameters are the neural network weights $\theta$ and the noise terms $\sigma \in \mathbb{R}^1$ and $\rho \in \mathbb{R}^1$. Further details are in Appendix \ref{appendix_rnn}.

\subsection{Evaluation}

We use hold-out log-likelihood to assess generalization to unseen data, a standard probabilistic approach in ML. The models were trained with 15 different random seed initializations to ensure the differences in the results were due to our approach as opposed to a quirk of a particular random seed. This is used to generate 95\% confidence intervals. Likelihood is not easily interpretable nor the end-goal of operational climate models. Ultimately we want to use weather and climate models to make forecasts, and it is common to measure forecast skill with error and spread \cite{gan_hannah,leutbecher2008ensemble} so this is also done for evaluation.

In each case we compare to the baseline of no transfer learning i.e. training solely on $\mathbf{X}_t$. The goal of our evaluation is to compare our approach against the baseline; it is not to assess if we have created the best model for a particular system.

\section{Results and Discussion}

We demonstrate our approach on three chaotic dynamical systems: (1) Kuramoto-Sivashinsky (KS), (2) Brusselator, and (3) Lorenz 96 (L96). These systems have been used extensively in emulation studies \cite{gan_hannah,vlachas2022multiscale,hannah_bespoke,crommelin2008subgrid,kwasniok2012data,rasp2020,pathak2017using,raissi2018hidden} and are well-suited here as we can mimic the separation of low- and high-resolution data. This is as follows: for the KS, $\mathbf{X}_t$ is a temporal and spatial averaging of $\mathbf{Y}_t$ across 5 time-steps and 5 spatial steps; for the Brusselator, it is as for the KS but the averaging is over 5 time-steps and 8 spatial steps; and for the L96, the model itself makes a separation between low- and high-resolution variables. Further details on each system are in Appendix \ref{appendix_systems}. 

For all three systems, the initial high-resolution training lasted 20 epochs. The number of epochs for the subsequent low-resolution fine-tuning was selected to ensure the models were trained enough to allow convergence if possible. This was 200 (KS), 400 (Brusselator) and 250 (L96). The training data consisted of sequences with the following number of points: 10,000 (KS), 600 (Brusselator) and 400 (L96). For each of the 15 randomly initialized models, early stopping was used during training on a validation sequence of 10,000 points.

\subsection{Hold-out Log-Likelihood}

The hold-out log-likelihood is shown  in Table \ref{tab:log_lik}. In all cases, a 30,000-length sequence is used as the hold-out set. A buffer between the training, validation and hold-sets is kept to ensure temporal independence. 

Our transfer learning approach improves on the baseline in all cases. Moreover, it gives far tighter confidence intervals over the average hold-out log-likelihoods. This suggests the approach helps better optima be reached in parameter space, improves generalisation, and reduces sensitivity to the initial random seed. Practically, the latter means reduced training costs as you do not need to run many randomly initialized training runs to account for the random seed's effect. 

The improved generalisation performance of our approach is also seen in what happens when early stopping is removed. Without it, our approach continues to fare well --- with continued training, the validation loss is stable; however, the no-TL approach begins to overfit and early stopping is required to select a reasonable model. 

\begin{table}[b]
  \caption{Log-likelihoods on hold-out set. Higher is better. `Max' is the hold-out log-likelihood of the model with the greatest validation log-likelihood. `Average' is the average of hold-out log-likelihoods for the 15 initializations. A 95\% confidence interval is included.}
  \label{tab:log_lik}
  \centering
   \begin{tabular}{@{}lllll@{}}
\toprule
\multirow{2}{*}{System} & \multicolumn{2}{l}{Transfer Learning}        & \multicolumn{2}{l}{No Transfer Learning} \\
                        & Max             & Average                    & Max           & Average                  \\ \midrule
KS                      & \textbf{93.09}  & \textbf{91.73 $\pm$0.62}   & 71.52         & 37.74 $\pm$16.50         \\
Brusselator             & \textbf{452.27} & \textbf{384.64 $\pm$27.41} & -1433.95      & -555.40 $\pm$486.39      \\
L96                     & \textbf{13.09}  & \textbf{11.51 $\pm$ 0.64}  & 6.78          & 4.37 $\pm$ 1.07          \\ \bottomrule
\end{tabular} 
    
    \end{table}

\subsection{Forecasting Skill}

\begin{figure}
  \centering
  \includegraphics[height=6cm]{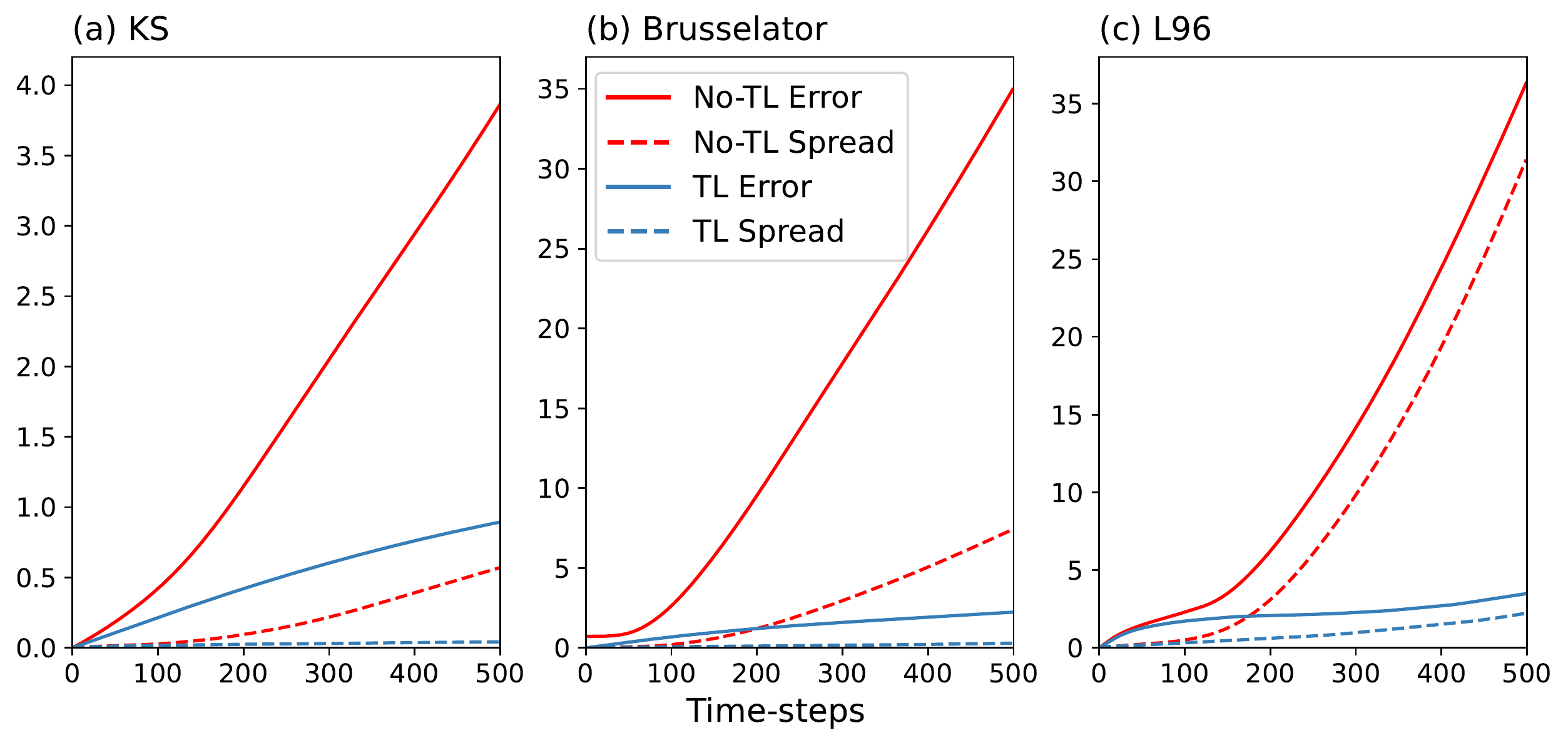}
  \caption{Forecast error and spread. In all cases the TL approach has far lower error. }
  \label{fig}
\end{figure}

Forecasts were made using the `Max' model in Table \ref{tab:log_lik} in the following way. $M=500$ initial conditions were randomly selected from data separate to the training set, and an ensemble of $N=40$ forecasts each 500 time-steps long were produced from each initial condition. Figure \ref{fig} shows (1) the error (intuitively, how far the ensemble mean is from the truth),  $\sqrt{\frac{1}{Md}\sum_{m=1}^{M}\sum_{i=1}^{d}\big(X_{m,i}^{O} (t)-\overline{X_{m,i}^{\text{sample}}(t)}\big)^2}$, where $X_{m,i}^{O} (t)$ is the $i^{\text{th}}$ dimension of the observed state at time $t$ for the $m^{\text{th}}$ initial condition, and $\overline{X_{m,i}^{\text{sample}}(t)}$ is the ensemble mean forecast at time $t$; and (2) the spread of the ensemble members, $\sqrt{\frac{1}{MNd}\sum_{m=1}^{M}\sum_{n=1}^{N}\sum_{i=1}^{d}\big(X_{m,n,i}(t)-\overline{X_{m,i}^{\text{sample}}(t)}\big)^2}$. 

Our TL approach results in significantly lower errors across the shown systems. In a perfectly reliable forecast, for large $N$ and $M$ the error should also be equal to the spread \cite{leutbecher2008ensemble}.The spreads are not well-matched in either approach though. 

\subsection{Effect of Training Data Size and Sensitivity to Hyperparameters}

As the size of the training set increases, the usefulness of the high-resolution data decreases. This is consistent with the TL literature. We found that in these systems, $\frac{\textrm{(\# neural network parameters)}^{0.1} d^{0.5} 10^4}{\textrm{(training sequence length)}^{1.5} } \lesssim 1$ (where $\mathbf{X}_t \in \mathbb{R}^d$) was a reasonable indicator for our approach providing less benefit. Importantly, as the training data increases, our approach did not worsen results compared to the baseline, instead just giving similar performance. 

Our approach is a form of regularization and should be seen as a complement to other forms such as dropout. Its usefulness depends on the need for regularization. We found that for a range of model complexities and data sizes it remained successful. However, if sufficient regularization was already provided by a simple enough network architecture, our approach provided no further benefit. Similarly, there were cases where increased regularization (more dropout; using our approach) for a particular model complexity could not prevent overfitting. In such cases, more data is often required.

\section{Conclusion}

By setting up the problem as a transfer learning task, we have shown how to learn from all high-resolution data to create low-resolution emulators. Our approach is a two-step process. It performs particularly well in data-scarce scenarios, acting as a regularizer.

We also tried multi-task learning, where fine-tuning is done on the low- and high-resolution data simultaneously, but it provided no benefit here. However, there may be other set-ups where it does provide value. 

There are potentially many helpful auxiliary tasks to use. For example, in operational forecasting models we might have high-resolution data only for specific spatial or temporal regions. Our approach is one way to learn from this. Looking ahead, this now needs testing in such operational models.






\begin{ack}
R.P. was funded by the Engineering and Physical Sciences Research Council [grant number  EP/S022961/1]. D.J.W. was funded by the University of Cambridge.

\end{ack}

\bibliographystyle{unsrt}
\small
\bibliography{bibliography.bib}

\normalsize
\appendix
\section{RNN Architecture}
\label{appendix_rnn}

\begin{figure}
  \centering
  \includegraphics[width=20cm,trim = 0cm 4cm 0cm 0cm]{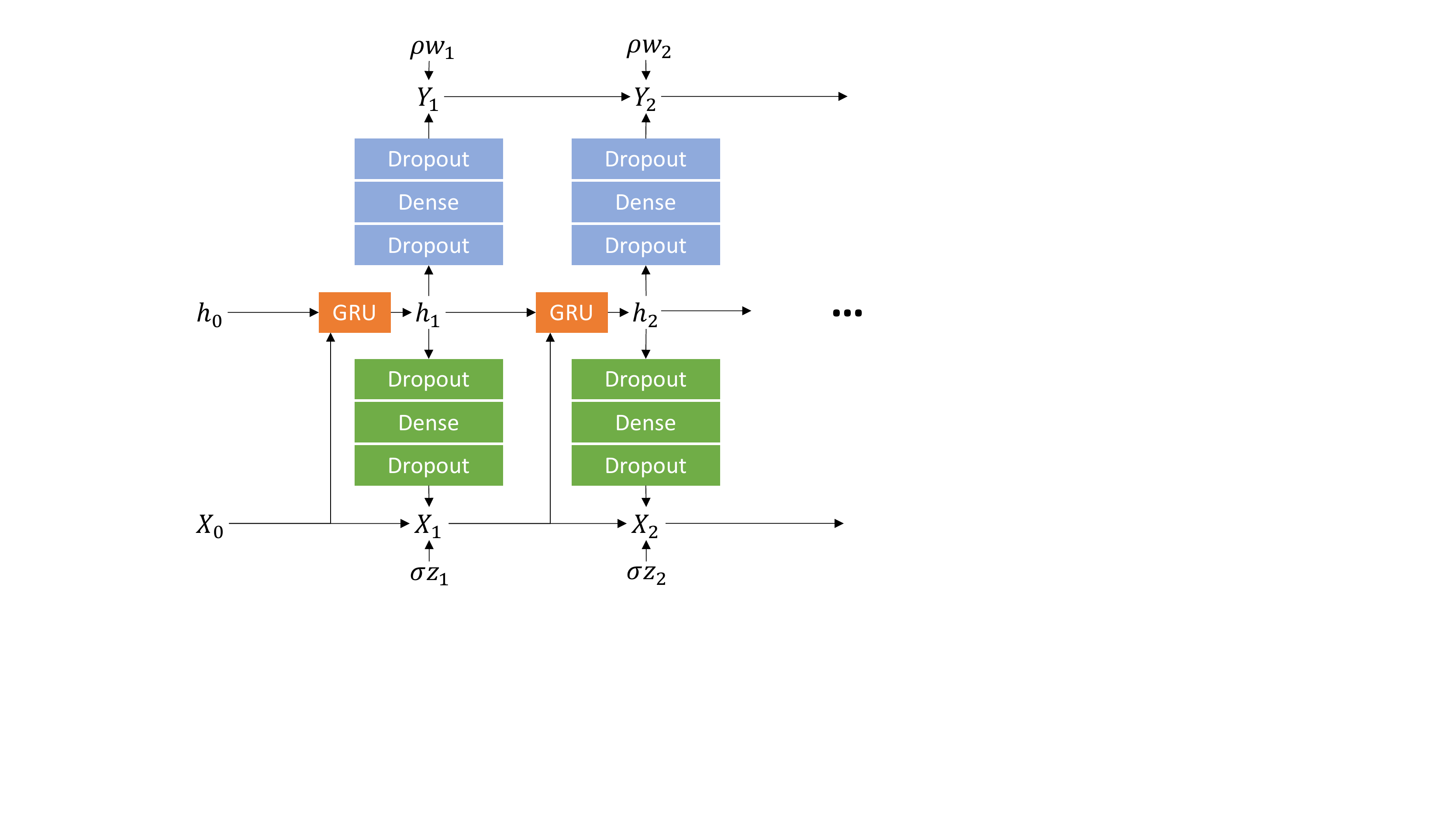}
  \caption{RNN graphical model showing how each $\mathbf{X}_{t}$ is generated. $\mathbf{z}_t$ and $\mathbf{w}_t$ are exogenous noise. $\mathbf{X}_{t}$ is not a function of $\mathbf{Y}_{t-1}$ so it is not necessary to track $\mathbf{Y}_{t}$ at simulation time.}
  \label{fig:rnn}
\end{figure}

In all cases, the RNN architecture takes the form as shown in the generative model in Figure \ref{fig:rnn}. There is a single GRU layer and it represents the function $f_\theta$ from $\mathbf{h}_{t+1} = f_\theta(\mathbf{h}_t,\mathbf{X}_t)$. The neural network layers mapping $\mathbf{h}_t$ to $\mathbf{Y}_t$ represent $j_\theta$ in equation \ref{eq:rnn2}, and those mapping $\mathbf{h}_t$ to $\mathbf{X}_t$ represent $g_\theta$ in equation \ref{eq:rnn1}. The dropout rate is 0.3 in all cases. All models were trained using truncated back propagation through time on sequences of length 100 time steps with a batch size of 32 using Adam \cite{kingma2014adam}. The number of nodes in each layer and the learning rates used are given in Table \ref{tab:hyperparameters}.

\begin{table}[]
  \caption{Hyperparameters used for each experiment.}
  \label{tab:hyperparameters}
  \centering
\begin{tabular}{@{}ccccc@{}}
\toprule
Experiment  & GRU units & Dense units for $\mathbf{X}_t$ & Dense units for $\mathbf{Y}_t$  & Learning rate \\ \midrule
KS          & 8         & 8                            & 16                                    & 0.001         \\
Brusselator & 8         & 64                           & 64                                    & 0.0003        \\
L96         & 32        & 32                           & 4                                     & 0.001        
\end{tabular}
\end{table}
\section{Details on Dynamical Systems}
\label{appendix_systems}

\subsection{Kuramoto-Sivashinsky}

The Kuramoto-Sivashinsky equation \cite{kuramoto1978diffusion,sivashinsky1977nonlinear} models diffusive-thermal instabilities in a laminar flame front. The one-dimensional KS equation is given by the PDE
\begin{equation}
    \frac{\partial{u}}{\partial{t}} = -\nu \frac{\partial^4{u}}{\partial{x^4}} - \frac{\partial^2{u}}{\partial{x^2}} - u \frac{\partial{u}}{\partial{x}}
    \label{eq:ks}
\end{equation}
on the domain $\Omega = [0,L]$ with periodic boundary conditions $u(0,t) = u(L,t)$ and $\nu = 1$. The case $L=22$ is used here as it results in a chaotic attractor. Equation \ref{eq:ks} is discretized with a grid of 100 points and solved with a time-step of $\delta t = 0.00005$ using the \textit{py-pde} package \cite{py-pde}. The data are subsampled to $\Delta t  = 0.002$ to create $\mathbf{Y}_t$. The first 10,000 steps are discarded. 

\subsection{Brusselator}

The Brusselator \cite{prigogine1968symmetry} is a model for an autocatalytic chemical reaction. The two-dimensional Brusselator equation is given by the PDEs
\begin{align} 
\frac{\partial{u}}{\partial{t}} &= D_{0} \nabla^2u + a - (1+b)u + vu^2 \label{eq:b1} \\ 
\frac{\partial{v}}{\partial{t}} &= D_{1} \nabla^2v + bu - vu^2 \label{eq:b2}
\end{align}
on the domain $\Omega = [[0,64],[0,64]]$ with $D_{0}=1,D_{1}=0.1,a=1$ and $b=3$. The parameters lead to an unstable regime. Here, $D_{0}=1$ and $D_{1}=0.1$ are diffusivities and $a$ and $b$ are related to reaction rates. Equations \ref{eq:b1}--\ref{eq:b2} are discretized on a unit grid and solved with a time-step of $\delta t = 0.0002$ using the \textit{py-pde} package \cite{py-pde}. The data are subsampled to $\Delta t  = 0.002$ to create $\mathbf{Y}_t$. The first 10,000 steps are discarded. 

\subsection{Lorenz 96}

The Lorenz 96 is a model for atmospheric circulation \cite{lorenz1996predictability}. We use the two-tier version which consists of two scales of variables: a large, low-frequency variable, $X_k$, coupled to small, high-frequency variables, $Y_{j,k}$. These evolve as follows
\begin{align}
    \frac{dX_k}{dt} &= -X_{k-1}(X_{k-2} - X_{k+1})- X_{k}+ F - \frac{hc}{b} \sum_{j=J(k-1)+1}^{kJ} Y_j  \label{eq:l1} \\
    \frac{dY_{j,k}}{dt} &= -cbY_{j+1,k}(Y_{j+2,k} - Y_{j-1,k})  - \, cY_{j,k}  - \frac{hc}{b} X_k  \label{eq:l2}
\end{align}
where in our experiments, the number of $X_k$ variables is $K=8$, and the number of $Y_{j,k}$ variables per $X_{k}$ is $J=32$. The value of the constants are set to $h=1,\:b=10$ and $\:c=10$. These indicate that the fast variable evolves ten times faster than the slow variable and has one-tenth the amplitude. The forcing term, $F$, is set to $F=20$. Equations \ref{eq:l1}--\ref{eq:l2} are solved using a fourth-order Runge-Kutta scheme with a time-step of $\delta t = 0.001$. The data are subsampled to $\Delta t  = 0.005$ to create $\mathbf{X}_t$ and $\mathbf{Y}_t$. The first 10,000 steps are discarded. 

\end{document}